\documentclass[10pt,twocolumn,letterpaper]{article}
\usepackage{cvpr}
\usepackage{graphicx}
\usepackage{times}
\usepackage{helvet,bbm}
\usepackage{courier}
\usepackage{multirow}
\cvprfinalcopy
\begin{document}
\title{Visual Learning of Arithmetic Operations}
\author{Yedid Hoshen ~~~~~~~~~~~~~ Shmuel Peleg\\School of Computer Science and Engineering\\The Hebrew University of Jerusalem\\Jerusalem, Israel}
\maketitle
\begin{abstract}
\begin{quote}
A simple Neural Network model is presented for end-to-end visual learning of arithmetic operations from pictures of numbers. The input consists of two pictures, each showing a 7-digit number. The output, also a picture, displays the number showing the result of an arithmetic operation (e.g., addition or subtraction) on the two input numbers. The concepts of a number, or of an operator, are not explicitly introduced. This indicates that addition is a simple cognitive task, which can be learned visually using a very small number of neurons.

Other operations, e.g., multiplication, were not learnable using this architecture. Some tasks were not learnable end-to-end (e.g., addition with Roman numerals), but were easily learnable once broken into two separate sub-tasks: a perceptual \textit{Character Recognition} and cognitive \textit{Arithmetic} sub-tasks. This indicates that while some tasks may be easily learnable end-to-end, other may need to be broken into sub-tasks. 
\end{quote}
\end{abstract}

\section{Introduction} 

\noindent Visual learning of arithmetic operations is naturally broken into two sub-tasks: A perceptual sub-task of optical character recognition (OCR) and a cognitive sub-task of learning arithmetic. A common approach in such cases is to learn each sub-task separately. Examples of popular perceptual sub-tasks in other domains include object recognition and segmentation. Cognitive sub-tasks include language modeling and translation. 

With the progress of deep neural networks it has become possible to learn complete tasks end-to-end. Systems now exist for end-to-end training of image to sentence generation \cite{vinyals2014show} and speech to sentence generation \cite{hannun2014deepspeech}. But end-to-end learning may introduce an extra difficulty: sub-tasks do not have unique training data, but depend on the results of other sub-tasks. 

\begin{figure*}[t]
\begin{tabular}{lccc}
 & Example A& Example B& Failure Example \\
\begin{tabular}[b]{l}
\vspace{28pt}
Input Picture 1\\
\vspace{26pt}
Input Picture 2\\
Network Output\\
\vspace{16pt}
Picture\\
Ground Truth\\
\vspace{2pt}
Picture
\end{tabular} & \includegraphics[width=0.20\textwidth]{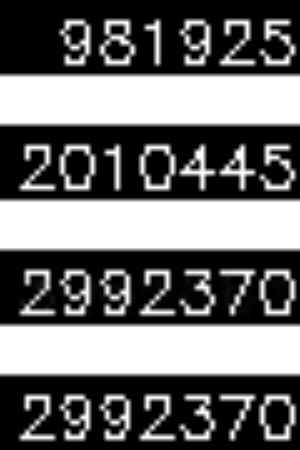} & \includegraphics[width=0.20\textwidth]{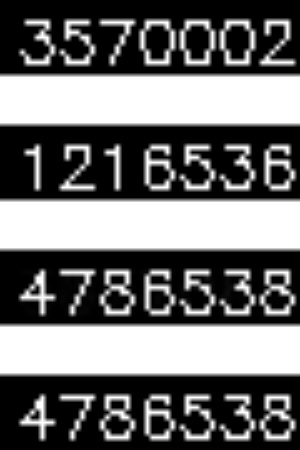} & \includegraphics[width=0.20\textwidth]{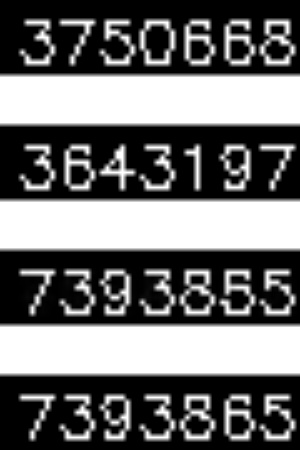}
\end{tabular}
\caption{\label{fig:clean}Input and output examples from our neural network trained for addition. The first two examples show a typical correct response. The last example shows a rare failure case. }
\end{figure*}

We examine end-to-end learning from a neural network perspective as a model for perception and cognition: performing arithmetic operations (e.g., addition) for visual input and visual output. Both input and output examples of the network are pictures (as in Fig.~\ref{fig:clean}). For each training example we give the student (the network) two input pictures, each showing a 7 digit integer number written in a standard font. The target output is also a picture, displaying the sum of the two input numbers.

In order to succeed at this task, the network is required to implicitly be able to learn the arithmetic operation without being taught the meaning of numbers. This can be seen as similar to teaching arithmetic to a person with whom we do not possess a common language.

We model the learning process as a feed-forward artificial neural network \cite{bishop1995neural,hinton2006reducing}. The input to the network are pictures of numbers, and the output is also a picture (of the sum of the input numbers). The network is trained on a sufficient number of examples, which are only a tiny fraction of all possible inputs. After training, given pictures of two previously unseen numbers, the network generates the picture displaying their sum. It has therefore learned the concept of numbers without direct supervision and also learned the addition operation. 

Although initially a surprising result, we present an analysis of visual learning of addition and demonstrate that it is realizable using simple neural network architectures. Other arithmetic operations such as subtraction are also shown to be learnable with similar networks. Multiplication, however, was not learned successfully under the same setting. It is shown that the multiplication sub-task is more difficult to realize than addition under such architecture. Interestingly, for addition with Roman numerals both the OCR and the arithmetic sub-tasks are shown to be realizable, but the end-to-end training of the task fails. This demonstrates the extra difficultly of end-to-end training.    

Our results suggest that some mathematical concepts are learnable purely from vision. An exciting possible implication is that some arithmetic concepts can be taught visually across different cultures. It has also been shown that end-to-end learning fails for some tasks, even though their sub-tasks can be learned easily. This work deals with arithmetic tasks, and future research is required to characterize what other non-visual sub-tasks can be learned visually e.g., by video frame prediction.   

\section{Arithmetic as Neural Frame Prediction}

In this section we describe a visual protocol for learning arithmetic by image prediction. This is done by training an artificial neural network with input and output examples.  

\subsection{Learning Arithmetic from Visual Examples}
\label{subsec:protocol}

Our protocol for visual learning of arithmetic is based on image prediction. Given two input pictures $F_1,F_2$, target picture $E$ is the correct prediction. The learner is required to predict the output picture, and the predicted picture is denoted $P$. The prediction loss is evaluated by the sum of square differences (SSD) between the pixel intensities of the predicted picture $P$ and the target picture $E$.
 
The input integers are randomly selected in a pre-specified range (for addition we use the range of [0,4999999]), and are written on the input pictures. The result of the arithmetic operation on the input numbers (e.g., their sum) is written on the target output picture $E$. The numbers were written on the pictures using a standard font, and were always placed at the same image position. See Fig.~\ref{fig:clean} for examples. 

Learning consists of training the network with $N$ such input/output examples (we use $N = 150,000$).

\subsection{Network Architecture}

In this section we present a method to test the feasibility of learning arithmetic given the protocol presented in Sec.~\ref{subsec:protocol}. Our simple but powerful learner is a feed-forward fully-connected artificial neural network as shown in Fig.~\ref{fig:net_arch}.  

The network consists of an input layer of dimensions $F_x {\times} F_y {\times} 2$ where $F_x$ and $F_y$ are the dimensions of the 2 input pictures. We used $F_x {\times} F_y = 15 {\times} 60$ unless specified otherwise. The network has three hidden layers each with 256 nodes with ReLU activation functions ($max(0,x)$) and an output layer (of the same height and width as the input pictures) with sigmoid activation. All nodes between adjacent layers were fully connected. An $L_2$ loss function is used to score the difference between the predicted picture and the expected picture. The network is trained via mini-batch stochastic gradient descent using the backpropogation algorithm.

\section{Experiments}

The objective of this paper is to examine if arithmetic operations can be learned end-to-end using purely visual information. To this end several experiments were carried out:

\subsection{Experimental Procedure}
\label{subsec:proc}

Using the protocol from Sec.~\ref{subsec:protocol} we generated 2 input pictures per example, each showing a single integer number. The numbers were randomly generated from a pre-specified range as detailed below.  The output pictures were created similarly, displaying the result of the arithmetic operation on the input.

The following arithmetic operations were examined:
\begin{itemize}
\item Addition: Sum of two 7 digit numbers, each in the range [0,4999999].
\item Subtraction: Difference between two 7 digit numbers in the range [0,9999999]. The first number was chosen to be larger or equal to the second number to ensure a positive result.
\item Multiplication: Product of two numbers, each in the range [0,3160].
\item Addition of Roman Numerals: Sum of two numbers in the range [0,4999999]. Both input and output were written in Roman numerals (IVXLCDM and another 7 numerals we "invented" from 5000 to 5000000). The longest number 9,999,999 was 35 numerals long. The medieval notation (IV instead of IIII) was not used.
\end{itemize}

For each experiment, 150,000 input/output pairs were randomly generated for training and 30,000 pairs were randomly created for testing. The proportion of numbers used for training is a very small fraction of all possible combinations.

We have also examined robustness to image noise of the addition experiment. Both input and output pictures were corrupted with a strong additive Gaussian noise.

\begin{figure}
\centering
\includegraphics[width=0.47\textwidth]{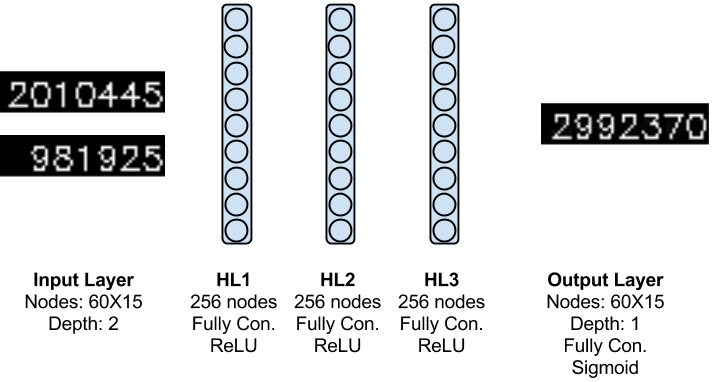}
\caption{\label{fig:net_arch}A diagram showing the construction of a neural network with 3 hidden layers able to preform addition using visual data. Two pictures are used as input and one picture as output. The network is fully connected and uses ReLU units in the hidden layers and sigmoid in the output layer. The hidden layers have 256 units each.}
\end{figure}

A feed-forward artificial network was trained with the architecture described in Fig.~\ref{fig:net_arch}. The network was trained using mini-batch stochastic gradient descent with learning rate 0.1, momentum 0.9 and mini-batch size was 256. 50 epochs of training were carried out. The network was implemented using the Caffe package \cite{jia2014caffe}.

\subsection{Results}
\label{subsec:res}
The correctness of the test set was measured using an OCR software (Tesseract \cite{smith2007overview} ) which was applied to the output pictures. The OCR results were compared to the desired output, and the percentage of incorrect digits was computed.
The effectiveness of the neural network approach has been tested on the following operations. 

\textit{Addition:} Three results from the test set are shown in Fig.~\ref{fig:clean}. The input and the output numbers were not included in the training set. The examples qualitatively demonstrate the effectiveness of the network at learning addition from purely visual information. Quantitatively, the network has been able to learn addition with great accuracy, with incorrect digit prediction rate being only 1.9\%.

\begin{figure*}
\begin{center}
\begin{tabular}{rccc}
 & Subtraction& Multiplication& Noisy Addition \\
\begin{tabular}[b]{l}
\vspace{28pt}
Input Picture 1\\
\vspace{26pt}
Input Picture 2\\
Network Output\\
\vspace{16pt}
Picture\\
Ground Truth\\
\vspace{2pt}
Picture
\end{tabular} & \includegraphics[width=0.20\textwidth]{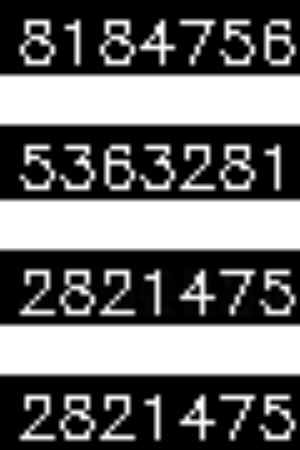} & \includegraphics[width=0.20\textwidth]{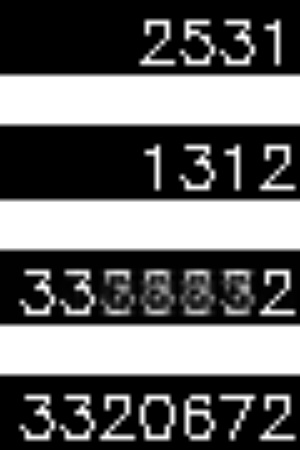} & \includegraphics[width=0.20\textwidth]{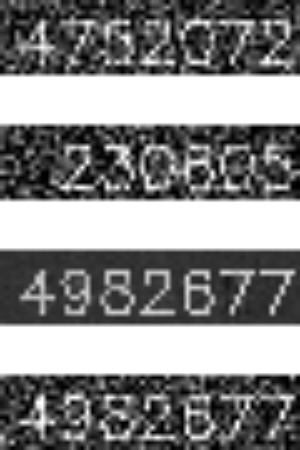}
\end{tabular}
\end{center}
\caption{\label{fig:more_ops}Examples of the performance of our network on subtraction, multiplication, and addition with noisy pictures. The network performs well on subtraction and is insensitive to additive noise. It performs poorly on multiplication. Note that the bottom right image is not the ground truth image, but an example of the type of training output images used in the Noisy Addition scenario.}
\end{figure*}

\textit{Subtraction:} We trained a neural network having identical architecture to the network used for addition. Subtraction of a small number from a larger one was found to be of comparable difficulty to addition. The predicted digit error rate was around 3.2\% which is comparable to addition.   

\textit{Multiplication:} This task was found to be a much more challenging operation for a feed-forward Neural Network. The data for this experiment consisted of two input pictures with 4-digit integers, resulting in an output picture with 7 digit number, and the network used was similar to the one used for addition. As theoretical work (e.g., \cite{franco1998solving}) has shown that multiplication of binary numbers may require two more layers than their addition, we experimented with adding more hidden layers. The network, even with 5 hidden layers, did not perform well on this task, giving very large train and test errors. An example input/output pair can be seen in Fig.\ref{fig:more_ops}. It can be seen that the least significant digit and two most significant digits were predicted correctly, as enumeration of the different possibilities is feasible, but the network was uncertain about the central 4 digits. The predicted digit error rate was as high as 71\%, and the OCR engine was often unable to read numbers that had several blurry (uncertain) digits.  

\textit{Addition of Roman numerals:} It has been hypothesized by Marr \cite{marr1982vision} and others (see \cite{schlimm2008modeling}) that arithmetic using Roman numerals can be more challenging than using Arabic numerals. We have repeated the addition experiment with all numbers written as Roman numerals, which can be up to 35 digits long. As is demonstrated quantitatively in Tab.~\ref{tab:dig_err} the network was not able to predict the output frame in Roman numeral basis. This suggests that end-to-end visual learning of addition in Roman numeral basis is more challenging, in agreement with Marr's hypothesis. We further analyze this result in Sec.~\ref{sec:disc}.        

\textit{Addition with Noisy Pictures:} In one experiment we added a strong Gaussian noise ($\sigma{=}0.3$) to all input and output pictures, as can be seen in Fig.\ref{fig:more_ops}. The network achieved very good performance on this task, giving output pictures that display the correct result, which are also clean from noise. Failures can occur when the input digits are almost illegible. In such cases the network generated a "probabilistic" output digit displaying a mixture of two digits. Mixture of digits caused problems to our verification using an OCR, reporting 9.8\% digit error rate whereas human inspection obtained only 3.2\% error rate. See Fig.\ref{fig:noisy_calc} for further details.


\begin{table}[bt]
\centering
\begin{tabular}{|l|c|c|c|c|} \hline
\multirow{2}{*}{Operation} & \multicolumn{2}{|c|}{Pictures} & \multicolumn{2}{|c|}{1-hot Vectors}\\ \cline{2-5}
 & \parbox[c]{1.0cm}{\raggedright No. Layers} & \% Error & \parbox[c]{1.0cm}{\raggedright No. Layers} & \% Error \\ 
\hline
Add  & 3 & 1.9\% & 1 & 1.7\% \\ \hline
Subtract  & 3 & 3.2\% & 1 & 2.1\% \\ \hline
Multiply & 5 & 71.5\% & 3 & 37.6\% \\ \hline
Roman & & & &  \\ Addition & 5 & 74.3 \% & 3 & 0.7 \% \\ \hline
\end{tabular}\\~\\

\caption{The digit prediction error rates for end-to-end training on pictures, and for the stripped 1-hot representation described in Sec.~\ref{sec:disc}. For the purpose of error computation, the digits in the output predicted images were found using OCR. Addition and subtraction are always accurate. The network was not able to learn multiplication. Although Roman numeral addition failed using the picture prediction network, it was learned successfully for 1-hot vectors. }
\label{tab:dig_err}

\end{table}

\section{Previous Work}
\label{sec:prev}

Theoretical characterization of the operations learnable by neural networks is an established area of research. A pioneering paper presented by \cite{hajnal1987threshold} used threshold circuits as a model for neural network capacity. A line of papers (e.g., \cite{hofmeister1991some,siu1993depth,franco1998solving}) established the feasibility of the implementation of several arithmetic operations on binary numbers. Recently \cite{graves2014neural} has addressed implementing Universal Turing Machines using neural networks. Most theoretical work in the field used binary representation of numbers, and did not address arithmetic operations in decimal form. Notably, a general result (see \cite{shalev2014understanding}), shows that operations implementable by Turing machine in time $T(n)$ can be implemented by a neural network of $O(T(n))$ layers and with $O(T(n)^2)$ nodes. It has sample complexity $O(T(n)^2)$ but has no guarantees on training time. Research has also not dealt with visual learning.

Hypotheses about the difference in difficulty of learning arithmetic using decimal vs. Roman representations was made by Marr \cite{marr1982vision} and others. see \cite{schlimm2008modeling} for a review and algorithms for Roman numeral addition and multiplication.

Optical Character Recognition (OCR) \cite{le1990handwritten,jaderberg2014deep} is a well studied field. In this work we only deal with a very simple OCR scenario, dealing with  more complex characters and backgrounds is out of scope of this work.

Learning to execute Python code (including arithmetic) from textual data has been shown to be possible using LSTMs by Zaremba and Sutskever \cite{zaremba2014learning}. Adding two MNIST digits randomly located in a blank picture has been performed by Ba et al. \cite{ba2014multiple}. In \cite{zaremba2014learningmath}, Recurrent Neural Networks (RNNs) were used for algebraic expression simplification. These works, however, required a non-visual representation of a number either in the input or in the output. In this paper we show for the first time that end-to-end visual learning of arithmetic is possible.

End-to-end learning of Image-to-Sentence \cite{vinyals2014show} and of Speech-to-Sentence \cite{hannun2014deepspeech} has been described by multiple researchers. A recent related work by Vondrick et al. \cite{vondrick2015anticipating} successfully learned to predict the objects to appear in a future video frame from several previous frames. Our work can also be seen as frame prediction, requiring the network to implicitly understand the concepts driving the change between input and output frames. But our visual arithmetic is an easier task: easier to interpret and to analyze. The greater simplicity of our task allows us to use raw frames rather than an intermediate representation as used in \cite{vondrick2015anticipating}.

\section{Discussion}
\label{sec:disc}

In this paper we have shown that feed-forward deep neural networks are able to learn certain arithmetic operations end-to-end by purely visual cues. Several other operations were not learned by the same architecture. In this section we give some intuition for the method the network employs to learn addition and subtraction, and the reasons why multiplication and Roman numerals were more challenging. A proof by construction of the capability of a shallow DNN (Deep Neural Network) to perform visual addition is presented in Sec.~\ref{sec:theory}.

\begin{figure}
\begin{center}
\includegraphics[width=0.20\textwidth]{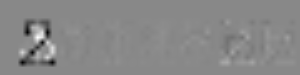}
\includegraphics[width=0.20\textwidth]{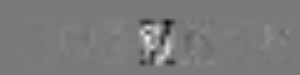}\\
(a) ~~~~~~~~~~~~~~~~~~~~~~ (b)\\
\includegraphics[width=0.20\textwidth]{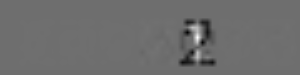}
\includegraphics[width=0.20\textwidth]{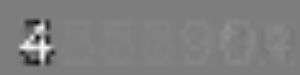}\\
(c) ~~~~~~~~~~~~~~~~~~~~~~ (d)\\
\end{center}

\caption{\label{fig:weights_top} (a-b) Examples of bottom layer weights for the first input picture. (a) recognizes '2' at the leftmost position, while (b) recognized '7' at the center position. ~~~~
(c-d) Examples of top layer weights. (c) outputs '1' at the second position, while (d) outputs '4' at the leftmost position. }
\end{figure}

When looking at the network weights for both addition and subtraction, we can see that each bottom hidden layer node is sensitive to a particular digit at a given position. Example bottom layer weights can be observed in Fig.~\ref{fig:weights_top}.a-b. The bottom hidden layer nodes therefore represent each of the two $M$-digit numbers as a vector of length $10 {\times} M$, each element representing the presence of digit $0-9$ in position $m \in [1,M]$. This representation of converting a variable with $D$ possible values (here 10) as $D$ binary variables all being $0$ apart from a single $1$ at the $d^{th}$ position is known as "1-hot". The top hidden layer contains a similar representation of the output number representing the presence of digit $0-9$ in position $m \in [1,M]$ with total size $10 {\times} M$. The task of the central hidden layers is mapping between the 1-hot representations of the input numbers (size $10 {\times} M {\times} 2$) and the 1-hot representation of the output number (size $10 {\times} M$).

The task is therefore split into 2 sub-tasks:
\begin{itemize}
\item Perception: learn to represent numbers as a set of 1-hot vectors. 
\item Cognition: map between the binary vectors as performed by the arithmetic operation.
\end{itemize}

Note that the second sub-task is different from arithmetic operations on binary numbers (and is often harder).

In order to evaluate the above sub-tasks separately, we repeated the experiments with the (input and output) data transformed to 1-hot representation, thereby bypassing the visual sub-task. We used the same architecture as in the end-to-end case, except that we removed the first and last hidden layers (that are used for detecting or drawing images of numbers at each location).

The results on the test sets measured as the percentage of wrong digits in the output number is presented in Tab.~\ref{tab:dig_err}. Addition and subtraction are both performed very accurately as in the visual case. The network was not able to learn multiplication due to the difficulty of the arithmetic sub-task, in line with the results of the visual case. This is also justified theoretically as (i) Binary multiplication was shown by previous papers \cite{siu1993depth,franco1998solving} to require deeper networks than binary addition. (ii)  The Turing Machine complexity of the basic multiplication algorithm (effective for short numbers) is $O(n^2)$ as opposed to $O(n)$ for decimal addition ($n$ is the number of digits). This means \cite{shalev2014understanding} that the operation is realizable only by a deeper ($O(n^2)$ vs. $O(n)$ layers) and larger network ($O(n^4)$ vs. $O(n^2)$ nodes).  

More interesting is the relative accuracy at which Roman numeral addition was performed, as opposed to the failure in the visual case. We believe this is due to the high number of digits for large numbers in Roman numerals (35 digits), which causes both input and output images to be very high dimensional. We hypothesize that convergence may be improved with preliminary unsupervised learning of the OCR tasks (i.e. teaching the network what numbers are by clustering). We conclude that Roman arithmetic can be learned by DNNs, but visual end-to-end learning is more challenging due to the difficulty of joint optimization with the OCR sub-task.

\begin{figure}
\begin{center}
\begin{tabular}{lc}
\begin{tabular}[b]{l}
\vspace{20pt}
Input Picture 1\\\\
\vspace{20pt}
Input Picture 2\\
Network Output\\
\vspace{5pt}
Picture
\end{tabular} & 
\includegraphics[width=0.20\textwidth]
{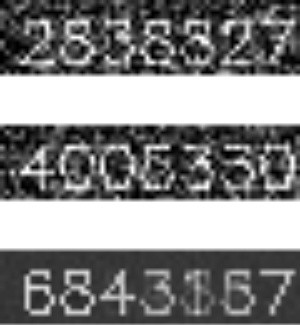}
\end{tabular}
\end{center}

\caption{\label{fig:noisy_calc} Probabilistic arithmetic for noisy pictures: The third digit from right in ``Input Picture 1" can be either 5 or 8. The corresponding output digit is a mixture of 1 and 8.}
\end{figure}

Visual learning when data were corrupted by strong noise was quite successful. In fact the concepts were learned well enough that the output pictures were denoised by the network. The performance on illegible digits is particularly interesting. We found that on corrupted digits that could possibly be read as multiple possibilities (In Fig.~\ref{fig:noisy_calc}, digits 8 or 5), the output digit also reflected this uncertainly, resulting in a mixture of the two possible outputs (In Fig.~\ref{fig:noisy_calc}, digits 1 or 8) with their respective probabilities. In other experiments (not shown) we have found that visual learning works for unary operations too (e.g., division by 2).

A significant difference between our model and the cognitive system is its invariance to a fixed permutation of the pixels. A human would struggle to learn from such images, but the artificial neural networks manages very well. This invariance can be broken by slight random displacement of the training data or by the introduction of a convolutional architecture.

Although Recurrent Neural Networks are generally better for learning algorithms (such as multiplication), we have chosen to use a fully connected architecture for ease of analysis. We hypothesize that better performance on multiplication can be obtained using an LSTM-RNN (Long Short Term Memory - Recurrent Neural Network) but we leave this investigation for future work.  

\begin{figure*}
\centering
\includegraphics[width=0.84\textwidth]{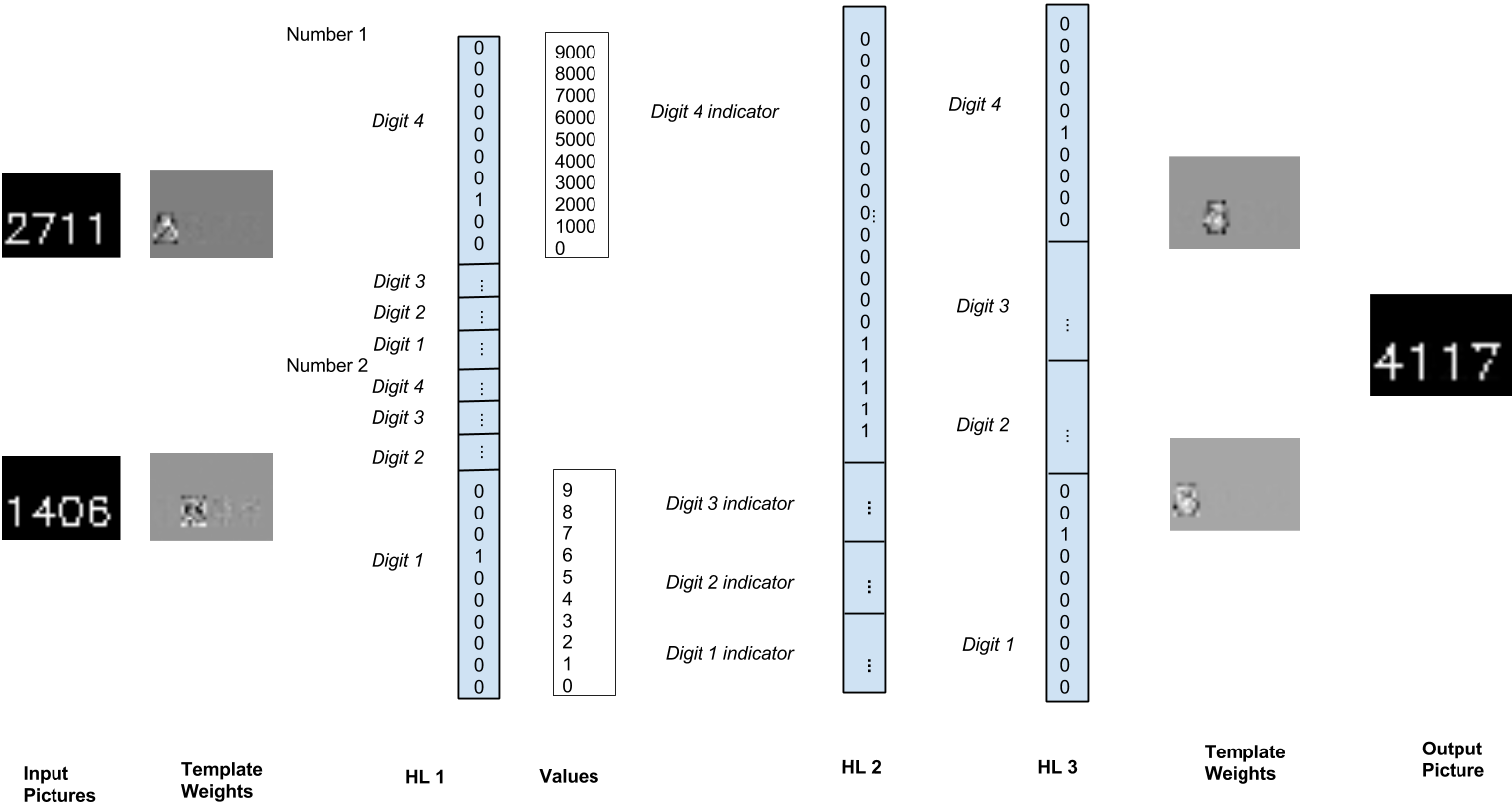}
\caption{\label{fig:construct} An illustration of the operation of a 3 hidden layer neural network able to perform addition using visual training. In this example the network handles only 4 digit numbers, but larger numbers are handled similarly with a linear increase in the number of nodes. i) The pictures are first projected onto a binary vector $HL1$ indicating if digit $n$ is present at position $m$ in each of the numbers. ii) In HL2 we compute indicator variables $v^m_i$ for each digit $1..M$ and threshold $i=0..19$. The variable is on if the summation result $\sum^m_{j=1}{(d1^m+d2^m) {\times} 10^j}$ exceeds threshold $i {\times} 10^m$.  iii) In the final hidden layer we calculate if a template is displayed by observing if the indicator variable corresponding to its digit and position is on but the following indicator variable is off. The templates are then projected to the output layer.  }
\end{figure*}

\section{Feasibility of a Visual Addition Network}
\label{sec:theory}

In this section we provide a feasibility proof by construction of a neural network architecture that can learn addition from visual data end-to-end. The construction of the network is illustrated in Fig.~\ref{fig:construct}. 

We rely on logic gates for simplicity. A logic gate can be implemented to an arbitrary accuracy by a single sigmoid or by a linear combination of 2 ReLU units $\Theta(x > 0) = (ReLU(x+\delta) - ReLU(x))/\delta$. Although our reported results were obtained using a network utilizing ReLU units, we have also tested our network with ReLU units replaced by sigmoid units obtaining similar results but much slower convergence. Logic gates are therefore a sufficiently good model of our network.

An input example is shown in Fig.~\ref{fig:clean}. The first layer of the network is a dimensionality reduction layer. We choose weights that correspond to the set of filters containing each digit n ($n \in 0..9$) at each position $m$. Our experimental network in fact chooses more complex filters usually concentrated between similar digits to increase accuracy of digit detection (see Fig.~\ref{fig:construct} for examples). We construct $10 {\times} M {\times} 2$ nodes in the HL1 layer indicating if each of the templates is triggered. Each first hidden layer node responds to a specific template, for example ${T2}^n_m$ corresponds to the template detecting if the digit $n$ is present at the $m^{th}$ position in picture 2. It has value 1 if a template appears and 0 if it does not. Similarly the output layer is represented as a set of templates each corresponding to a digit (0..9) at a given position ($1..M$).

It is worth noticing that given two digits $d1^m$ and $d2^m$ at the $m^{th}$ position in numbers 1 and 2 respectively, the $m^{th}$ digit in the output $d_o^m$ can be either $(d^m_1 + d^m_2) mod 10$ or $(d^m_1 + d^m_2 + 1) mod 10$. For each pair of digits, the arithmetic problem is to choose the correct result from the possible two. 

In $HL2$ we compute an indicator function for each digit $m$, where node $v^m_i$ is on when the sum of digits $d_1$ and $d_2$ and the possible increment from previous digits is larger than its threshold $i$ ($i \in 0..19$). This is formulated as 
\begin{equation}
v^m_i =  \mathbbm{1}_{\sum^m_{j=1}{(d1^m+d2^m)*10^j} >= i \times 10^m}
\label{eqn:hl2}
\end{equation}
It is easily implemented for each node $v^m_i$ with weights from $HL1$ nodes $T1^m_n$ and $T2^m_n$ with values $n*10^j$ for $j \in 1..m$, $n \in 0..9$ and threshold $i {\times} 10^m$. For later convenience we denote $v^m_{20} = 0$.

In $HL3$, output template $o^m_n$ corresponding to the digit $n$ at position $m$ is turned on if in $HL2$ indicator $v^m_n = 1$ or $v^m_{n+10} = 1$ while $v^m_{n+1} = 0$ or $v^m_{n+11} = 0$ respectively. This corresponds to the cases where the summation result of the numbers up to digit $m$ is $n \times 10^m \leq result < (n+1) \times 10^m$ or $(n+10) \times 10^m \leq result < (n+11) \times 10^m$. The equation is therefore:
\begin{equation}
o^m_n =  \mathbbm{1}_{v^m_{n} - v^m_{n+1} + v^m_{n+10} - v^m_{n+11} > 0}
\label{eqn:hl3}
\end{equation}

Finally the values are projected onto the output picture using the corresponding digit templates. 

By end-to-end training of the network with a sufficient number of examples the network can arrive at the above weights (although it is by no means guaranteed to), and in practice good performance is achieved. End-to-end training from visual data is therefore theoretically shown and experimentally demonstrated to be able to learn addition with little guidance. This is a powerful paradigm that can generalize to visual learning of non-visual concepts that are not easily directly communicated to the learner.

\section{Conclusions}
\label{sec:conc}

We have examined the capacity of neural networks for learning arithmetic operations from pictures, using a visual end-to-end learning protocol. Our neural network was able to learn addition and subtraction, and was robust to strong image noise. The concept of numbers was not explicitly used. We have shown that the network was not able to learn some other operations such as multiplication, and visual addition using Roman numerals. For the latter we have shown that although all sub-tasks are easily learned, the end-to-end task is not.

In order to better understand the capabilities of the network, a theoretical analysis was presented showing how a network capable of performing visual addition may be constructed. This theoretical framework can help determine if a new arithmetic operation is learnable using a feed-forward DNN architecture. We note that such analysis is quite restrictive, and hypothesize that experimental confirmation of the end-to-end learnability of complex tasks will often result in surprising findings.

Although this work dealt primarily with arithmetic operations, the same approach can be used for general cognitive sub-task learning using frame prediction. The sub-tasks need not be restricted to the field of arithmetic, and can include more general concepts such as association. Generating data for the cognitive sub-task in not trivial, but generating visual examples is easy, e.g., by predicting future frames in video.

While our experiments use two input pictures and one output picture, the protocol can be generalized for more complex operations involving more input and output pictures. For learning non-arithmetic concepts, the pictures may contain other objects beside numbers.\\ 

\noindent
{\bf Acknowledgments.}
This research was supported by Intel-ICRC and by the Israel Science Foundation. The authors thank T. Poggio, S. Shalev-Shwartz, Y. Weiss, and L. Wolf for fruitful discussions.



\bibliographystyle{ieee}
\bibliography{finalref}

\end{document}